\documentclass[10pt,journal]{IEEEtranTCOM}

\usepackage[english]{babel}
\usepackage[usenames]{color}
\usepackage[cp1250]{inputenc}
\usepackage{amsfonts}
\usepackage{amsthm}
\usepackage{graphicx}
\usepackage{epsfig}
\usepackage{mathrsfs}
\usepackage{amsmath}
\usepackage{algorithm}
\usepackage{algorithmic}
\usepackage{hyperref}
\usepackage[table]{xcolor}

\theoremstyle{plain}

\begin{document}
\newcommand{\bea}{\begin{eqnarray}}
\newcommand{\eea}{\end{eqnarray}}
\newcommand{\be}{\begin{equation}}
\newcommand{\ee}{\end{equation}}
\newcommand{\beas}{\begin{eqnarray*}}
\newcommand{\eeas}{\end{eqnarray*}}
\newcommand{\bs}{\backslash}
\newcommand{\bc}{\begin{center}}
\newcommand{\ec}{\end{center}}
\def\SC {\mathscr{C}}

\title{Fast optimization of common basis for matrix set\\ through Common Singular Value Decomposition}
\author{\IEEEauthorblockN{Jarek Duda}\\
\IEEEauthorblockA{Jagiellonian University,
Golebia 24, 31-007 Krakow, Poland,
Email: \emph{dudajar@gmail.com}}}
\maketitle

\begin{abstract}
SVD (singular value decomposition) is one of the basic tools of machine learning, allowing to optimize basis for a given matrix. However, sometimes we have a set of matrices $\{A_k\}_k$ instead, and would like to optimize a single common basis for them: find orthogonal matrices $U$, $V$, such that $\{U^T A_k V\}$ set of matrices is somehow simpler. For example DCT-II is orthonormal basis of functions commonly used in image/video compression - as discussed here, this kind of basis can be quickly automatically optimized for a given dataset. While also discussed gradient descent optimization might be computationally costly, there is proposed CSVD (common SVD): fast general approach based on SVD. Specifically, we choose $U$ as built of eigenvectors of $\sum_i (w_k)^q (A_k A_k^T)^p$ and $V$ of $\sum_k (w_k)^q (A_k^T A_k)^p$, where $w_k$ are their weights, $p,q>0$ are some chosen powers e.g. 1/2, optionally with normalization e.g. $A \to A - rc^T$ where $r_i=\sum_j A_{ij}, c_j =\sum_i A_{ij}$.
\end{abstract}
\textbf{Keywords:} machine learning, statistics, feature extraction, matrix set, function basis, SVD, PCA, DCT, data compression, hierarchical correlation analysis
\section{Introduction}
SVD (singular value decomposition)~\cite{svd}, PCA (principal component analysis)~\cite{pca}, Karhunen-Lo\`{e}ve transform~\cite{kar}, CCA (canonical correlation analysis)~\cite{cca}, MCA (multiple correspondence analysis) are related basic tools of statistics and machine learning, e.g. allowing to optimize a basis. They are focused on optimization of basis a single matrix, bringing a natural question of extensions to optimization of single basis simultaneously for multiple matrices.

PCA can be seen as applying SVD to covariance matrix $C_{XX}=E[(X-E[X])(X-E[X])^T]$, in CCA cross-covariance matrix $E[\bar{X}\bar{Y}^T]$ for whitened variables $\bar{X}=C_{XX}^{-1/2}(X-E[x])$. If splitting the sample into subsamples in $(w_k)$ proportion, the final e.g. covariance matrix is weighted average of  matrices for these subsamples with $(w_k)$ weights.

Hence optimization of common basis of set of (cross-) covariance matrices $(A_k)_{k=1..K}$ with $(w_k)$ weights, can be performed with SVD of $\sum_k w_k A_k$. The question is what if they are a different type of matrices?

\begin{figure}[t!]
    \centering
        \includegraphics[width=8.5cm]{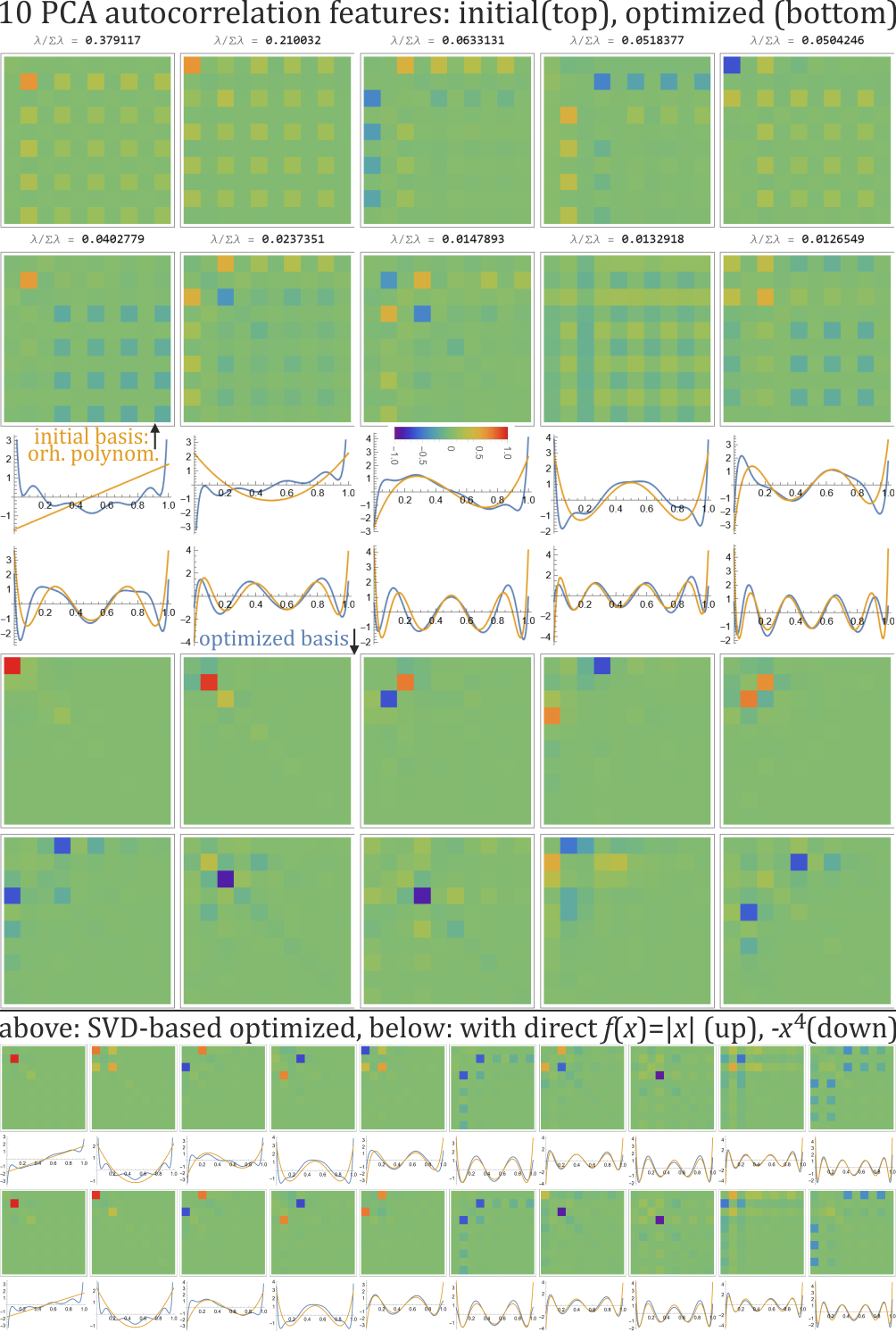}
        \caption{Authors' original motivation for HCR (hierarchical correlation analysis): multi-feature autocorrelation analysis of bearing time series (data source: \url{https://engineering.case.edu/bearingdatacenter/}) as in \cite{hcr}. Time series was first normalized to nearly uniform distribution on $[0,1]$, then for pairs of points shifted by $\tau$ lag, there was MSE fitted joint distribution as linear combination $\sum_{ij=1..10} a^\tau_{ij} f_i(y) f_j(z)$ with basis $f_i$ of orthonormal polynomials (orange plots for $i=1,\ldots,10$). This way for each lag $\tau$ we get 100 coefficients, hence there was performed PCA to find dominating features as their linear combinations - presented in top diagrams above with written contributions to variance. We can see patterns suggesting that this arbitrarily chosen basis could be optimized. Blue plots show such optimized common basis using proposed CSVD, at the bottom there are shown these PCA features in this new basis. We can see they are nearly canonical: single value in the diagonal, or symmetric/anti-symmetric pairs outside diagonal - suggesting to replace complex PCA features with such simple canonical ones in optimized basis, which should be more universal. }
        \label{basis}
\end{figure}

 \begin{figure*}[t!]
    \centering
        \includegraphics[width=17 cm]{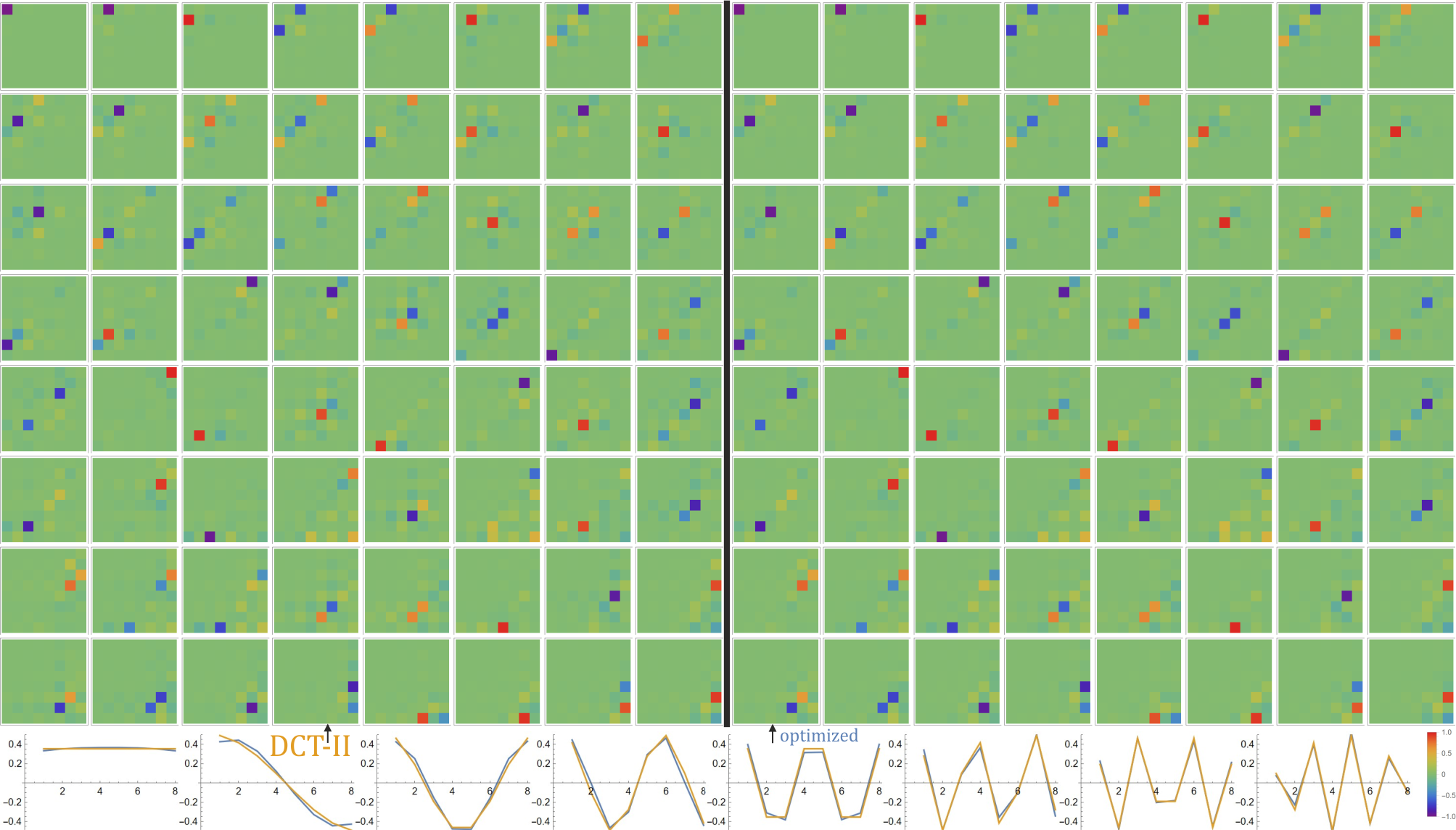}
        \caption{Comparison of DCT-II basis popular in image/video compression (left, orange plots), with basis optimized by proposed CSVD (right, blue plots). For dataset of 48 grayscale 8bit images $512\times 512$ (source: \url{http://decsai.ugr.es/cvg/CG/base.htm}), there were extracted $48\cdot 64\cdot 64$ blocks of $8\times 8$ pixels, for which there was performed PCA - getting 64 eigenvectors, presented in DCT-II basis (top left). For these 64 matrices $8\times 8$ there was also applied discussed CSVD optimization (with eigenvalues as weights), getting basis plotted in blue, in which these 64 matrices are presented top right. We can see this optimization has automatically lead to nearly identical basis as DCT-II, confirming this choice and e.g. allowing dedicated local optimizations for some specific datasets. Presented 64 matrices ideally should have only a single nonzero values - such basis is used in practice, but this is not exactly true here (allowing for optimizations), the optimized ones on the right are visually slightly smoother.    }
        \label{dct}
\end{figure*}
Practical application example is optimization of basis $(f_i(x))_i$ for which we want to use product basis $(f_i(x) f_j(y))_{ij}$. It is convenient to work with orthonormal basis, but there is large freedom of choosing a specific one. For example in lossy image/video compression like JPEG~\cite{jpeg} there was finally chosen DCT-II~\cite{dct}: $\cos(\pi (n+1/2)k/N)$. But there are 4 types of DCT (discrete cosine transform), also bases using sines instead, orthonormal polynomials, etc. - bringing question why to choose DCT-II? Maybe we could choose a better basis optimized for a given dataset? Split data using bases optimized for each cluster? To optimize it based on dataset, we can e.g. take $8\times 8$ pixel blocks, perform PCA getting 64 eigenvectors giving $8\times 8$ matrices, and eigenvalues suggesting their importance: weights, then optimize common basis for these 64 matrices - done in Fig. \ref{dct}, confirming that DCT-II is a very good choice, and allowing further e.g. local optimizations.

The direct author's motivation was very similar, presented in Fig. \ref{basis} - also optimization of basis (starting with orthogonal polynomials), which product basis we want to optimize based on PCA from dataset - this time representing joint probability distribution in autocorrelation analysis. This application allows to optimize more universal features - instead of dataset dependent PCA features, optimize basis to further use simple universal canonical features.

The article further discusses direct optimization of chosen evaluation function - through gradient descent in space of orthogonal matrices, however, it is extremely costly and often non-unique. There is also proposed CSVD approach family (common SVD): which is SVD for weighted sum of $(A_k A_k^T)^p$, with freedom of choosing the power $p$ (or a different function), adding normalization, etc.

\section{Direct optimization by gradient descend}
The choice of $U,V$ orthogonal matrices, transforms the given $\{A_k\}$ matrices into $\{U^T A_k V\}$ with coefficients:
\be c_{ijk} = (U^T A_k V)_{ij} = (U_{\bullet i})^T A_k V_{\bullet j} \ee
for $U_{\bullet i}=(U_{ji})_j$, $V_{\bullet j}=(V_{ij})_i$ denoting their columns as vectors of the optimized orthonormal bases.

We need to define the evaluation function on $(c_{ijk})$. Its simple form is coordinate-wise:
\be\min_{UV:UU^T=I_n, VV^T=I_m} g\qquad\textrm{for}\qquad g=\sum_{ijk} w_k f(c_{ijk}) \label{opt}\ee
Generally it might be also worth to consider matrix-wise evaluation functions: $\sum_k w_k f((c_{ijk})_{ij})$ using more complex matrix norms/functions, or even completely general $f((w_k)_k,(c_{ijk})_{ijk})$ e.g. optimizing maximal norm over the matrix set.

Optimization over orthogonal matrices, after a step $U\to U(1+\epsilon G)$ needs to remain orthogonal: $I =UU^T \to U(I+\epsilon G)(I+\epsilon G^T) U^T $, zeroing of $\epsilon$ term says generator $G$ has to be anti-symmetric: $G^T = -G$.

Hence the $G,H$ generators have correspondingly $n(n-1)/2$, $m(m-1)/2$ coefficients, which can be referred with $G_{ab}=-G_{ba}, H_{ab}=-H_{ba}$ indexes for $a < b$.

This way e.g. $G_{ab}$ adds $U_{\bullet b} \to U_{\bullet b} + G_{ab} U_{\bullet a}$, and also subtracts $U_{\bullet a}^T \to U_{\bullet a}^T - G_{ab} U_{\bullet b}^T$, leading to derivatives for $U\to U(1+\epsilon G)$, $V\to V(1+\epsilon H)$:
$$ \frac{\partial g}{\partial G_{ab}}=\sum_{jk} w_k \left(f'(c_{bjk}) c_{ajk} -f'(c_{ajk}) c_{bjk}  \right) $$
$$ \frac{\partial g}{\partial H_{ab}}=\sum_{ik} w_k \left(f'(c_{ibk}) c_{iak} -f'(c_{iak}) c_{ibk}  \right) $$
For square matrix and symmetric $U^T=V$ case, we have only a single generator $G=H$ and derivative is sum of the two above. Having the gradient we can perform gradient descent step $U\to U(1-\epsilon G)$, $V\to V(1-\epsilon H)$ , which needs to interleaved with some orthonormalization steps e.g. Gram-Schmidt. The choice of $\epsilon>0$ step is difficult, there can be tested a few and chosen the one with best evaluation.

In bottom of Fig. \ref{basis} there are shown effects of such direct coordinate-wise optimization for two coordinate-wise:
\begin{itemize}
  \item $f(x)=|x|$: L1-like evaluation function hoping to get sparse representation like in so called lasso, and for 
  \item $f(x)=-x^4$: sums of 2nd powers are rotation invariant, maximizing 4th powers wants to get high contrasts.
\end{itemize}
It is quite costly numerically, for $10\times 10$ matrices already requiring thousands of gradient descent steps. It might be worth to expand to 2nd order methods.

Also the results are not that satisfactory - visually better looking and much less expensive computationally is the discussed next SVD-based optimization, which could be also used as initial point for further direct optimization.

 \begin{figure}[t!]
    \centering
        \includegraphics[width=8.5cm]{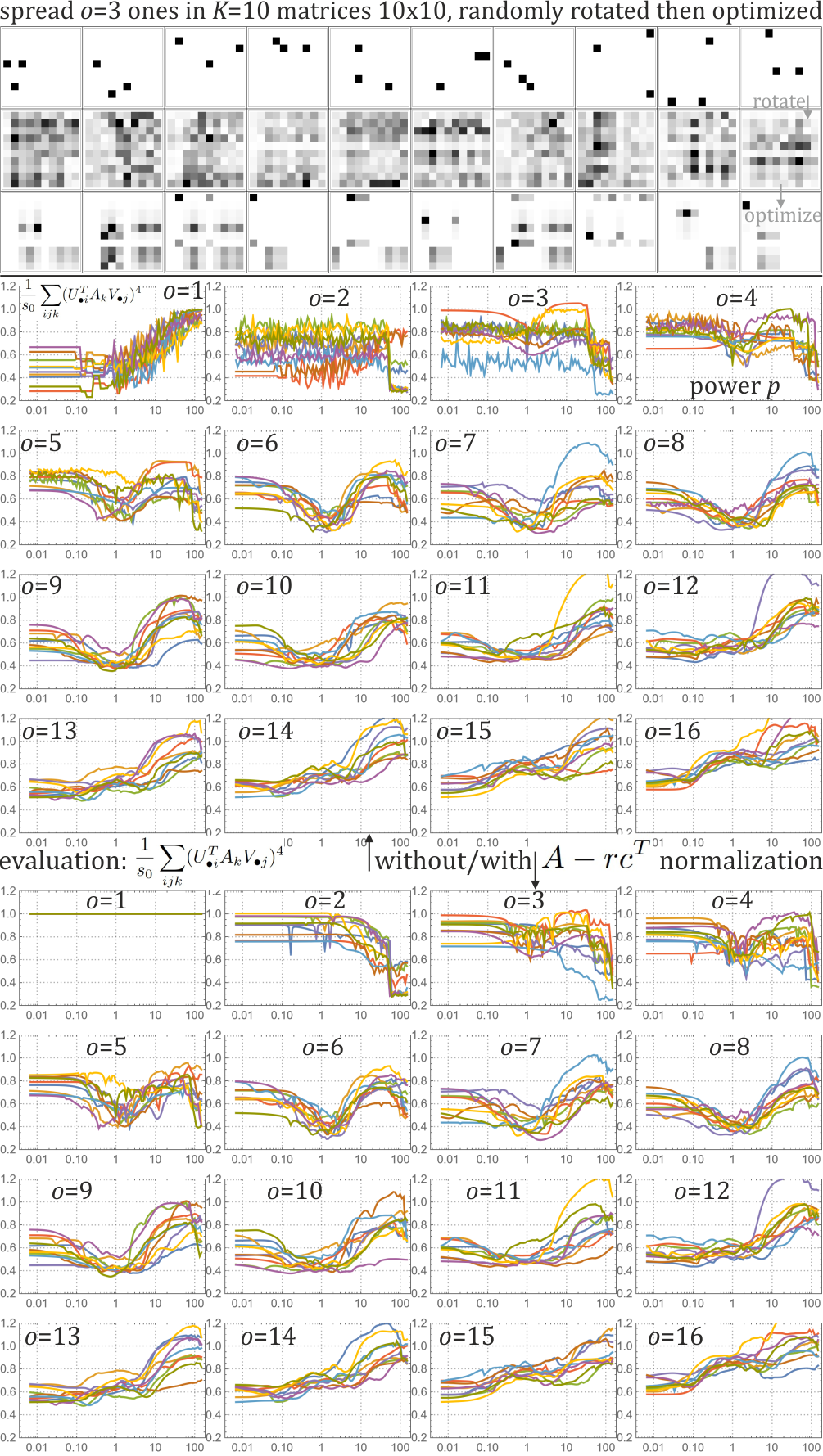}
        \caption{Test example of discussed CSVD as in example on top: we generate $\{M_k\}_{k=1..10}$ matrices $10\times 10$ by randomly distributing $o=3$ 1s (upper row, the rest are 0). Then randomly generate orthogonal $U$, $V$, and transform $A_k = U M_k V^T$ (central row). Then CSVD is used to find a common basis with constant weights and $p=1$ power (lower row). Below:   with (above)/without(below) $A\to A-rc^T$ normalization results of $\frac{1}{s_0} \sum_{ijk} (U_{\bullet i}^T A_k V_{\bullet j})^4$ evaluation (vertical axis), where $s_0$ is such sum for the original matrix $M_k$.  Horizontal axis shows the used power $p$, for various number of 1s $o$, each contains results of 10 experiments (generation of random $(M_k), U, V$). We can see that large powers often give high evaluation, normalization helps in low $o$ cases. }
        \label{test}
\end{figure}

\section{CSVD - common SVD optimization}
Returning to PCA analogy: $E[(X-E[X])(X-E[X])^T]$ covariance matrices for subsets of dataset in $(w_k)_k$ proportions, covaraince matrix for the entire dataset is their weighted average with $(w_k)$ weights - we can optimize the common basis through SVD of this weighted average.

Having general $(A_k)_k$ set of matrices of the same size $n\times m$ with $(w_k)_k$ weights, in some cases it might be worth to directly consider their weighted average $M=\sum_k w_k A_k$ and then perform SVD, what in practice can be done by calculating eigenvectors of $MM^T$ and of $M^T M$ (both symmetric, positive defined) and building $U,V$ orthogonal matrices from these eigenvectors as columns.

However, experimentally much better behavior is obtained by performing weighted average of $n\times n$ matrices $A_k A_k^T$ and $m\times m$ matrices $A_k^T A_k$. In covariance matrix analogy, this way imagining the matrices as $m$ size $n$ vector dataset, or the opposite. We can also include some functions before weighting to optimize for specific applications, e.g. powers:
\newline 

\noindent\textbf{Common SVD (CSVD)}: for $(A_k)_k$ real $n\times m$ matrices with $(w_k)_k$ weights, and some chosen $p,q>0$ powers: choose orthogonal $U$ as built of eigenvectors of $\sum_k (w_k)^q (A_k A_k^T)^p$, and $V$ of $\sum_k (w_k)^q (A_k^T A_k)^p$ (optionally $A_k$ normalized).
\subsection{Basic analysis}
To optimize a common basis, we should not fix the order of its vectors between the matrices. In contrast, SVD finds vectors sorted by singular values. To formulate SVD with permutation freedom, observe that while $\textrm{Tr}(MM^T)=\sum_{ij} (M_{ij})^2$ is unchanged under $M\to U^T M V $ transformation, SVD finds $U,V$ zeroing all non-diagonal coefficients of $U^T M V$ (leaving only singular values in diagonal), what allows to formulate it through optimization:

\noindent\textbf{Permutation-free SVD} of $M$ matrix - maximize:
\be \max_{\substack{U:UU^T=I\\V:VV^T=I}} \sum_i \left((U^T M V)_{ii}\right)^2
=\sum_{ij} (M_{ij})^2 \label{svd} = \textrm{Tr}(MM^T)\ee
For symmetric $M=M^T$ it can be simplified:
\be \max_{U:UU^T=I} \sum_i \left((U^T M U)_{ii}\right)^2 = \textrm{Tr}(M^2)=\sum_{i} (\lambda_i)^2 \label{svd1}\ee

Where $\lambda_i$ are the eigenvalues. In discussed CSVD, (\ref{svd1}) is applied to linear combination $\sum_k \bar{w} B_k$ for 
\be \bar{w}_k=(w_k)^q\qquad\qquad B_k = (A_k A_k^T)^p\ee
 or some more complex function e.g. containing normalization. Using (\ref{svd1}) we find $UU^T=I$ maximizing:
$$ \sum_i \left(U_{\bullet i}^T \left(\sum_k \bar{w}_k B_k\right) U_{\bullet i} \right)^2=$$
$$= \sum_i \left(\sum_k w_k U_{\bullet i}^T B_k U_{\bullet i} \right)^2=$$
\be \sum_{k,i} \bar{w}_k^2 U_{\bullet i}^T (B_k)^2 U_{\bullet i}+\sum_{k\neq k',i}\bar{w}_k \bar{w}_{k'} U_{\bullet i}^T B_k B_{k'} U_{\bullet i} \ee
The left hand side seems naturally weighted (\ref{svd1}), especially if choosing $p=q=1/2$. However, the right hand side part mixes between the matrices, what might be unwanted. Its contribution could be removed by using such found basis as initial step of further direct optimization.

We can also reduce this matrix mixing by \textbf{normalization} shifting them to be somehow localized around zero. It is done e.g. for covariance matrix by subtracting the means $(X\to X-E[X])$, in MCA by subtracting marginal distribution using both means:
\be A \to A- rc^T \qquad \textrm{for}\quad r_i=\sum_j A_{ij},\quad c_j =\sum_i A_{ij}\label{norm}\ee
individually for each matrix, maybe also applying whitening e.g. by multiplying by $C_{XX}^{-1/2}$. Tests of various approaches suggest (\ref{norm}) seems the most promising.

Figure \ref{test} contains experiments for 10 of $10\times 10$ matrices - first randomly filled with $o=1..16$ of 1s (the remaining positions are 0), then randomly chosen orthogonal matrices and applied to all, then used OSGD and evaluating $\sum_{ijk} (U_{\bullet i}^T A_k V_{\bullet j})^4$ divided by of the initial matrix, depending on power $p$ in horizontal axis. We can see e.g.  that (\ref{norm}) normalization has allowed for perfect choice of basis for $o=1$ and improved for small $o$.

It also shows that choice of power $p$ is quite complex and important, could be optimized for various applications. Large $p$ leaves in $B_k$ nearly only projection on the highest eigenvector, tiny $p$ deforms $B_k$ into nearly identity - both can reduce the mixing term, but also deform the optimized condition.
\section{Conclusions and further work}
Optimization of common basis for set of matrices can allow for nearly free optimization e.g. in data compression, feature extraction, statistical analysis. While it seems a difficult task and direct optimization is quite costly, turns out inexpensive SVD-based approach gives promising results.

One direction for further work is choosing better evaluation functions e.g. for direct optimization, or maybe with inexpensive direct solutions, approximations like SVD-based.

Another related direction is finding applications and optimizing for them, e.g. in data compression - there are now considered various transformations e.g. based on cosines, sines, asymmetric - we could optimize some better ones, also locally based on dataset, maybe use some model clustering~\cite{cluster}: split data into clusters with automatically optimized models.

\bibliographystyle{IEEEtran}
\bibliography{cites}
\end{document}